\documentclass[conference]{IEEEtran}
\IEEEoverridecommandlockouts
\usepackage{amsmath,amssymb,amsfonts}
\usepackage{algorithmic}
\usepackage{graphicx}
\usepackage{textcomp}
\usepackage{hyperref}
\usepackage{subfigure}
\usepackage{bbm}
\usepackage{comment}
\usepackage{svg}
\usepackage{xcolor}

\def\BibTeX{{\rm B\kern-.05em{\sc i\kern-.025em b}\kern-.08em
    T\kern-.1667em\lower.7ex\hbox{E}\kern-.125emX}}

\usepackage{xcolor}

\newcommand{\facets}{FACETS }
\newcommand{\FACETS}{\facets}

\usepackage[sorting=none]{biblatex}
\begin{document}
\title{Toward a Realistic Benchmark for Out-of-Distribution Detection
}

\author{\IEEEauthorblockN{Pietro Recalcati, Fabio Garcea, Luca Piano, Fabrizio Lamberti, Lia Morra}
\IEEEauthorblockA{\textit{Department of Control and Computer Engineering} \\
\textit{Politecnico di Torino}\\
Torino, Italy\\
\{name.surname\}@polito.it }}

\maketitle

\begin{abstract}
Deep neural networks are increasingly used in a wide range of technologies and services, but remain highly susceptible to out-of-distribution (OOD) samples, that is, drawn from a different distribution than the original training set. A common approach to address this issue is to endow deep neural networks with the ability to detect OOD samples. Several benchmarks have been proposed to design and validate OOD detection techniques. However, many of them are based on far-OOD samples drawn from very different distributions, and thus lack the complexity needed to capture the nuances of real-world scenarios. In this work, we introduce a comprehensive benchmark for OOD detection, based on ImageNet and Places365, that assigns individual classes as in-distribution or out-of-distribution depending on the semantic similarity  with the training set. Several techniques can be used to determine which classes should be considered in-distribution, yielding benchmarks with varying properties.  Experimental results on different OOD detection techniques show how their measured efficacy depends on the selected benchmark and how confidence-based techniques may outperform classifier-based ones on near-OOD samples. 
\end{abstract}

\begin{IEEEkeywords}
Out-of-Distribution Detection, Deep Learning, Convolutional Neural Networks, Open-World recognition
\end{IEEEkeywords}

\section{Introduction}
\label{sec:introduction}
Deep convolutional networks (CNNs) are powerful classifiers when tested on in-distribution (ID) images sampled from the same distribution the network was trained on. However, being trained under a closed-world assumption, they may fail by producing overconfident and wrong results when faced with out-of-distribution (OOD) samples, such as images belonging to classes previously unseen by the model. There is a strong interest in making CNN classifiers more robust by endowing them with the capability to separate samples drawn from a given distribution (also known as inliers, in-distribution or ID samples) from the others (also denoted as outliers, out-of-distribution, OOD, anomalies, novelties, or out-of-domain samples) \cite{salehi2021unified,yang2021generalized,amodei2016concrete,relu_nets_conf}. 

As a motivating example, let us consider the automatic tagging of images from social media platforms such as Facebook or Instagram, with applications in social sciences \cite{wilson2012review}, digital humanities \cite{karafillakis2021methods,santangelo2023}, marketing \cite{salminen2019machine}, etc. Researchers may want to exploit readily available, pre-trained models for automatic classification and tagging, which however entails operating under an open-world assumption. Hence, it is necessary to detect OOD samples that may lead to overconfident and wrong predictions. In the following, we will refer to the task of scene classification as our primary case study.

Several methods were proposed in the literature for OOD detection \cite{salehi2021unified}. However, their experimental comparison is complicated by the broad definition of OOD and the wide variety of settings under which they were tested. The performance of an OOD detector intrinsically depends on the experimental setting and its underlying assumptions. For instance, distance-based methods were shown to yield better performance than those based on prediction scores depending on whether the OOD samples are far away or close to the decision boundary between classes \cite{no_true_state_of_the_art}. 

One of the fundamental aspects that differentiate OOD benchmarks is the semantic and visual distance between ID and OOD samples. Many works in literature have drawn ID and/or OOD samples from small, low-resolution datasets such as CIFAR10, CIFAR100 or SVHN \cite{fort2021exploring,zhou2021learning,oodl,outlier_exposure, li2020background, liu2020energy, malinin2018predictive}. This choice is not representative of real applications, such as social media tagging, where the difference between ID and OOD samples is expected to be more nuanced and dependent on the underlying class semantics. Other authors have used inter- and intra-dataset comparison in order to construct more realistic benchmarks: for instance, in \cite{roady2019out}, Roady and colleagues proposed both inter-dataset comparison (e.g., using ImageNet as ID and Places365 as OOD or viceversa) and intra-dataset comparison (e.g., using a subset of ImageNet as ID and another subset as OOD). 

\begin{figure}
    \centering
    \includegraphics[width=\columnwidth]{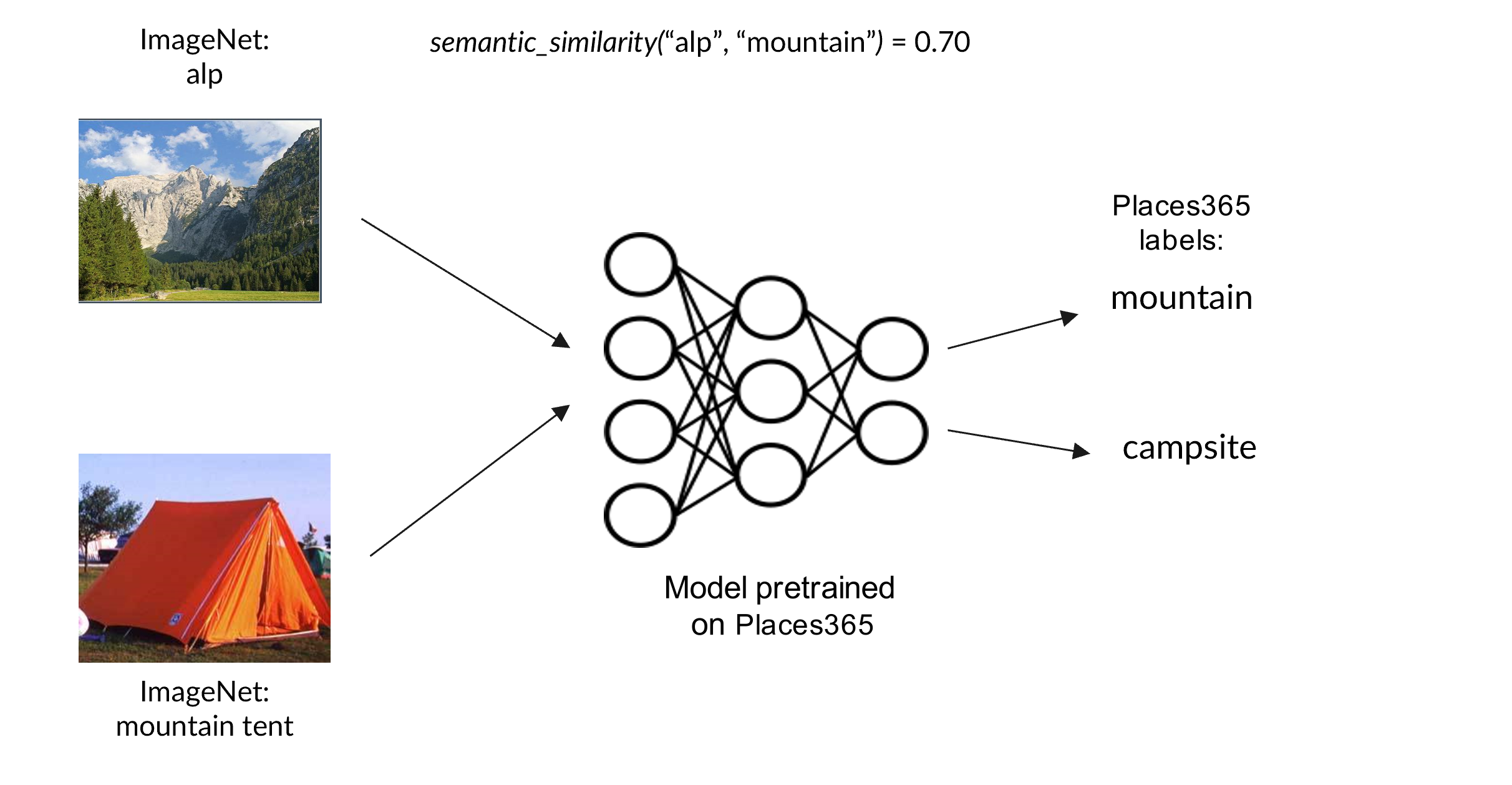}
    \caption{Examples of predictions generated by a classifier pretrained on Places365 (right) on samples drawn from the ImageNet dataset (left). The predicted classes, although different from the ImageNet labels, are highly semantically correlated. 
    }
    \label{fig:imagenetVSPlacesModel}
\end{figure}

Both options, however, have conceptual or practical drawbacks: inter-dataset comparison ignores or under-estimates semantic overlap between different datasets, whereas intra-dataset comparison modifies the training set and thus cannot be applied \textit{as is} to existing pre-trained models. A clarifying example is shown in \autoref{fig:imagenetVSPlacesModel}, in which two images from ImageNet belonging to classes \textit{alp} and \textit{mountain tent}, respectively, are classified by a scene classifier pre-trained on Places365 as \textit{mountain} and \textit{campsite}, respectively (\autoref{fig:imagenetVSPlacesModel}). Although the classes are distinct and would technically be considered OOD according to conventional inter-dataset approaches, the predictions would be still deemed acceptable from a practical standpoint by the end-user. Instead, for the intra-dataset scenario, consider the classes  \textit{dressing\_room}, \textit{beauty\_saloon}, and \textit{closet} from the Place365 dataset. These three classes may have overlapping features such as lighting, color scheme, and furniture arrangement, which can lead to confusion for the model during training and inference. As a result, the performance of the model on these classes may be lower than expected due to the limited ability to differentiate between them. Benchmarks for OOD detection should be aligned with the ultimate goal of rejecting unknown samples and avoiding high-confidence predictions, regardless of the specific dataset they are drawn from. Hence, we argue that the decision of whether a sample should be considered ID or OOD cannot be based on the source dataset alone, but rather should take into account the semantic content of the class/image.

In this paper, a benchmark for OOD detection is constructed based on Places365 (selected as ID) and ImageNet (from which OOD samples are drawn). ImageNet classes were sorted based on their semantic affinity with Places365 classes in order to classify them as either ID, near-OOD, or far-OOD. To this aim, different techniques, including automatic and manual tagging, were compared. Then, different methods for OOD detection were evaluated on the proposed benchmark, showing that more realistic benchmarks affect the ranking of OOD detection methods. The proposed datasets are available for download  \footnote{\url{https://huggingface.co/datasets/GrainsPolito/FACETS_Datasets}}.

The rest of the paper is organized as follows. A detailed examination of related work, including various techniques employed to detect OOD samples and relevant datasets used to test these methods is reported in Section~\ref{sec:related_work}. Then, Section~\ref{sec:methodology} delves into the problem definition for OOD detection.
Section~\ref{sec:datasets} illustrates the proposed datasets designed to test the efficacy of different OOD detection strategies under diverse conditions.
Section~\ref{sec:preliminary_analysis} presents some initial findings from our investigations.
Finally, Section~\ref{sec:conclusions}  showcases the outcome of our experimental study, offering valuable insights into strengths and weaknesses of different techniques.

\section{Related Work}
\label{sec:related_work}

\subsection{OOD Detection}
OOD detection has attracted much attention and several efforts have been made to solve this problem. While many OOD detection techniques are general and can be in principle applied to any type of machine learning model or classifier, we focus here in particular on the task of image classification. For other input data and machine learning models, the reader is referred to existing surveys and benchmarks \cite{jacob2021exathlon,schmidl2022anomaly}. 

Existing OOD detection methods generally fall into two broad categories: \textit{classifier-based} approaches and \textit{generative} techniques. While the latter explicitly model the target distribution to reject samples that are not consistent with it, the former attempt to increase the robustness of an existing model or pair it with an external classifier for OOD detection. Classifier-based methods are a large family of discriminative approaches.  Among those, \textit{confidence-based} techniques strive to generate a numerical score by directly exploiting the classifier predictions. The softmax and the logit functions are popular options for this metric. 

The seminal work by Hendrycks and Gimpel \cite{hendrycks2016baseline} introduced the concept of OOD detection and proposed a first benchmark and evaluation metrics. Despite prior interest in related problems throughout the 1990s, their work was the foundation for subsequent advancements in the field. Specifically, they employed the maximum softmax probability (MSP) as the scoring metric for detecting OOD samples. Liang \textit{et al.} proposed an enhanced variant of the MSP called ODIN (OOD detector for Neural Networks) \cite{liang2017enhancing}. ODIN was designed to increase the discriminatory power between ID and OOD samples by using a temperature scale factor (T), applied to the softmax output, and calculating the final score after perturbing the input image by a constant magnitude, established based on the loss function gradient. Generalized ODIN \cite{hsu2020generalized} further improves by removing explicit reliance on OOD data to determine suitable values for the temperature T and the perturbation magnitude. It achieves this objective by switching temperature scaling for a function of the input and setting the perturbation magnitude value that maximizes the softmax for ID samples. 

Conventional OOD identification algorithms face challenges when dealing with large semantic spaces, as pointed out by Huang and Li \cite{huang2021mos} when proposing the Minimum Others Score (MOS) method, which groups categories together to reduce complexity. As a result, the decision boundary between known and unknown samples becomes less challenging for the classifier assigned to each group, improving its OOD detection abilities. Replacing the softmax function with raw logits as the OOD detection score is another way of tackling the issue of a large semantic space.The method known as Maximum Logit Value (MLV) \cite{MLV} effectively reduces the dispersion of probability mass among classes with similar features. According to Bendale and Boult \cite{bendale2016towards}, OpenMAX could replace the softmax function for the purpose of OOD detection. The  approach leverages the intrinsic relationship among the activations of multiple classes, collectively denoted as an Activation Vector (AV). The position of a Mean Activation Vector (MAV) for each class in relation to the AV of an input image is then employed for OOD detection. More recently, Meinke and Hein \cite{meinke2020confidence} presented Certified Certain Uncertainty (CCU), which exhibits high predictive confidence across a wide range of input variations outside of the training distribution.

\subsection{OOD Datasets}
In previous studies, the most common benchmarking method used was to designate one dataset as an ID, while several additional datasets were designated as OOD. The choice of datasets is often similar among studies and driven by the ease of collecting or processing established research benchmarks. Therefore, this method facilitates comparison with other techniques that have been tested on the same benchmark. 

\begin{table}[t]
\caption[Usage statistics for the 10 most popular data sources used as either ID or OOD.]{Usage statistics for the 10 most popular data sources designated as either ID or OOD in experimental evaluation of OOD detection techniques.} 
\label{tab:ch1_ood_data}
\centering
\resizebox{\linewidth}{!}{
\begin{tabular}{cccccc}
\cline{2-6}
\multicolumn{1}{c|}{} &
  \multicolumn{1}{c|}{\textbf{CIFAR-10}} &
  \multicolumn{1}{c|}{\textbf{CIFAR-100}} &
  \multicolumn{1}{c|}{\textbf{SVHN}} &
  \multicolumn{1}{c|}{\textbf{MNIST}} &
  \multicolumn{1}{c|}{\textbf{TinyImageNet}} \\ \hline
\multicolumn{1}{|c|}{\textbf{Popularity}} &
  \multicolumn{1}{c|}{87\%} &
  \multicolumn{1}{c|}{57\%} &
  \multicolumn{1}{c|}{57\%} &
  \multicolumn{1}{c|}{56\%} &
  \multicolumn{1}{c|}{38\%} \\ \hline
\multicolumn{1}{|c|}{\textbf{ID}} &
  \multicolumn{1}{c|}{55} &
  \multicolumn{1}{c|}{26} &
  \multicolumn{1}{c|}{20} &
  \multicolumn{1}{c|}{32} &
  \multicolumn{1}{c|}{13} \\ \hline
\multicolumn{1}{|c|}{\textbf{OOD}} &
  \multicolumn{1}{c|}{39} &
  \multicolumn{1}{c|}{24} &
  \multicolumn{1}{c|}{34} &
  \multicolumn{1}{c|}{29} &
  \multicolumn{1}{c|}{21} \\ \hline
\multicolumn{1}{l}{} &
  \multicolumn{1}{l}{} &
  \multicolumn{1}{l}{} &
  \multicolumn{1}{l}{} &
  \multicolumn{1}{l}{} &
  \multicolumn{1}{l}{} \\ \cline{2-6} 
\multicolumn{1}{l|}{} &
  \multicolumn{1}{c|}{\textbf{LSUN}} &
  \multicolumn{1}{c|}{\textbf{Fashion-MNIST}} &
  \multicolumn{1}{c|}{\textbf{Gaussian Noise}} &
  \multicolumn{1}{c|}{\textbf{Textures}} &
  \multicolumn{1}{c|}{\textbf{Places365}} \\ \hline
\multicolumn{1}{|c|}{\textbf{Popularity}} &
  \multicolumn{1}{c|}{37\%} &
  \multicolumn{1}{c|}{19\%} &
  \multicolumn{1}{c|}{22\%} &
  \multicolumn{1}{c|}{17\%} &
  \multicolumn{1}{c|}{17\%} \\ \hline
\multicolumn{1}{|c|}{\textbf{ID}} &
  \multicolumn{1}{c|}{0} &
  \multicolumn{1}{c|}{9} &
  \multicolumn{1}{c|}{1} &
  \multicolumn{1}{c|}{0} &
  \multicolumn{1}{c|}{1} \\ \hline
\multicolumn{1}{|c|}{\textbf{OOD}} &
  \multicolumn{1}{c|}{23} &
  \multicolumn{1}{c|}{9} &
  \multicolumn{1}{c|}{14} &
  \multicolumn{1}{c|}{11} &
  \multicolumn{1}{c|}{11} \\ \hline
\end{tabular}}
\end{table}

To gather information on the prevalent options for ID and OOD datasets, we examined a collection of 63 research articles published from 2015 to 2021. We illustrate the 10 most frequently chosen datasets in \autoref{tab:ch1_ood_data}. The popularity of each dataset is indicated by the number of publications wherein the dataset has been employed as either ID or OOD data source. Toy datasets such as CIFAR-10 \cite{CIFAR}, CIFAR-100, MNIST \cite{MNIST}, and Fashion-MNIST \cite{Fashion-MNIST}, featuring a modest number of classes and low-resolution images, are quite popular as they enable rapid experimentation. These simple yet effective datasets continue to be favored by the community. CIFAR-10, in particular, features prominently in 87\% of the examined publications. This dataset comprises a vast collection of low-resolution images belonging to 10 distinct categories. Other widely employed datasets are  CIFAR-100 and SVHN, used in 57\% of publications, followed by MNIST.  The utilization of Fashion-MNIST \cite{Fashion-MNIST} as a substitute for MNIST has gained increasing popularity, accounting for 19\% of the examined publications. This dataset comprises 10 categories of clothing items and features 60,000 training low-resolution, grayscale images. Its creation aimed to offer a more demanding benchmark problem compared to MNIST while preserving similarities in sample and dataset sizes. 

TinyImageNet \cite{TinyImageNet} and LSUN \cite{LSUN} are used in less than half of the publications surveyed. With 200 and 10 classes, respectively, these datasets present an increasing complexity compared to the previous ones. Specifically, TinyImageNet is a reduced set of ImageNet of 100,000 images, whereas LSUN consists comprises images from 10 distinct scene categories.

The Gaussian Noise and Textures benchmarks \cite{cimpoi2014describing} account for only 22\% and 17\% of the papers examined. Typically, they are employed to assess the resilience of ML algorithms against noise or texture modifications. The former comprises images sourced from the ID datasets with varying degrees of Gaussian noise superimposed, whereas the latter includes 5,640 images representing 47 distinct texture classes, such as brick or grass. Finally, the Places365 dataset \cite{Places365}, which encompasses more than 1.8 million images across 365 diverse scenes, appears in just 17\% of the papers surveyed.

Interestingly, some datasets have been utilized more frequently as OOD sources rather than ID. For instance, datasets such as Gaussian Noise and Places365, which were employed as ID sources in just one publication each, are commonly utilized as OOD sources. Likewise, datasets such as Textures and LSUN, while never used as ID sources, were widely employed as OOD sources. 
These findings suggest that many researchers use distinct datasets as ID and OOD data.

There has been limited exploration in combining OOD detection techniques across diverse datasets. Recently, Roady et al. \cite{roady2019out} presented a benchmark suite that evaluates the capacity of OOD detection algorithms to scale. Specifically, they assessed the performance of OOD detection approaches using two widely-used image classification datasets, i.e., ImageNet-1K and Places-434. They created three separate OOD settings of varying difficulty. For the first one, labeled \textit{Noise}, synthetic images were generated from a Gaussian distribution ensuring that the resulting images reflected the normalization used for both the training and test images. The second one, called \textit{Inter-Dataset}, is of intermediate difficulty and involves testing each method's ability to detect OOD samples drawn from another large-scale dataset. The last one, \textit{Intra-Dataset}, is designed to identify novel classes within a specific dataset. The researchers used the same training set and models for all three settings, only varying the testing set. Ten thousand ID instances from each class in the validation set are randomly chosen to create the latter. Additionally, ten thousand outliers from the OOD classes within each dataset validation set are also selected. This approach however ignores or underestimate the semantic overlap between classes from both datasets, since it relies on exact name correspondence to determine overlapping classes. Conversely, the intra-dataset comparison has its shortcomings because it directly influences the training set and cannot be straightforwardly applied on available pre-trained models.
\section{Methodology}
\label{sec:methodology}

\subsection{Problem Definition}

Let $\mathcal{D}_{I}$ be the probability distribution of ID samples, including $K_{I}$ different classes, and $\mathcal{D}_{O} = \{\mathcal{D}_{O}^i\}_{i=1}^{\infty}$ the set of distributions of OOD samples. 
In the most general setting, at training time a dataset $\mathcal{T} = \mathcal{T}_{I} \cup \mathcal{T}_{O}$ is available, where 
$$\mathcal{T}_{I} = \{(x_{i}, y_{i}, z_{i})\}_{i=1}^{N_{I}},$$
$$ \text{with } x_{i} \sim \mathcal{D}_{I} \text{, } y_{i} \in \{1,\dots, K_{I}\} \text{, } z_{i} = 0$$
is the set of ID training samples and
$$\mathcal{T}_{O} = \{(x_{i}, y_{i}, z_{i})\}_{i=1}^{N_{O}},$$ $$ \text{} x_{i} \sim \mathcal{D}_{O}^{j} \in \mathcal{D}_{O} \text{, } y_{i} \in \{1,\dots, K_{O}^{j}\} \text{, } z_{i} > 0 \text{, } j \in \{1, \dots, J\}$$
is the set of OOD training points. In particular, each $x_{i}$ is a training image drawn from a distribution, $y_{i}$ is the corresponding class index, $K_{O}^{j}$ is the number of classes modeled by distribution $\mathcal{D}_{O}^{j}$, $J$ the number of OOD distributions represented in $\mathcal{T}_{O}$, and $z_{i}$ the ground truth label for OOD detection (henceforth referred to as \textit{OODness}) for the $i$-th sample, which is $0$ if the sample is ID and $1$ if the sample is OOD. Within this general formulation, several settings are possible. Some methods assume that OOD samples are not available at training time ($\mathcal{T}_{O} = \{\empty\}$). Other assumes that OOD samples are available, although not necessarily drawn from the same distribution/classes observed at inference time, which are generally unknown in real-world applications.

The classifier or machine learning model is then defined as a vector function $$f: \mathbb{R}^{d} \rightarrow \mathbb{R}^{K_{I}}, \mathbf{x} \mapsto \mathbf{l},$$
where $\mathbf{x}$ is a $d$-dimensional input and $\mathbf{l}$ the vector of class scores, referred to as \textit{logits}. The model $f$ is composed of $L$ concatenated layers, whose output $\mathbf{l^{i}}$ is calculated as $$\mathbf{l^{i}} = f^{i}(\mathbf{x}), i \in \{1, \dots, L\},$$
having defined $f^{i}$ as the composition of layers $1, \dots, i$. The whole classifier $f$ then coincides with $f^{L}$, and the logits vector $\mathbf{l}$ with $\mathbf{l^{L}}$. Class scores $\mathbf{l}$ are converted to class probabilities $\mathbf{p}$ via the application of the softmax function 
$$
p_{i} = S(\mathbf{l})_{i} = \frac{e^{\mathbf{l}_i}}{\sum_{j=1}^{K_{I}} e^{\mathbf{l}_{j}}} \approx \mathbb{P}[y=i|x=\textbf{x}],
$$
where $\mathbf{l} = f(\mathbf{x})$ and $y$ is the true label for input $\mathbf{x}$.
The predicted class $c$ is then chosen to be the one for which the assigned probability is the highest:
$c = \text{argmax}(\mathbf{p})$. The network is usually trained by minimizing the cross-entropy loss $\mathcal{L}_{CE}$ between the predicted probabilities and the one-hot-encoded vector $\mathbf{y}$ representing the ground truth.
The ability to reject undesired inputs is provided by an additional function $s$ that returns the OOD score $o = s(\mathbf{x})$ for sample $\mathbf{x}$. This score is then discretized by selecting a suitable threshold, in order to obtain a binary prediction as either ID or OOD sample.

One of the main differences among research works lays in the choice of the scoring function $s$, which can be obtained by extracting some internal network outputs, by means of an additional classifier or by explicit density estimation. The threshold $\theta$ is usually selected with the aim of optimizing a relevant metric for the specific task and dataset.

\subsection{Scoring Methods for OOD Detection}

Although OOD detection approaches that exploit OOD samples during training have shown to be effective, the wide range of potential unobserved distributions or classes constitute a fundamental limitation to their effectiveness. These techniques generally model OOD detection by using a predefined collection of OOD distributions, and increasing their diversity may not necessarily improve performance when exposed to new samples at inference time. In addition, post-hoc scoring methods, that do not require costly training procedures,  are advantageous when the classifier $f$ is pre-trained. We will thus here focus on OOD detection methods that do not require a large number of OOD samples at training time, and in which the use of OOD samples is mainly limited to parameter selection.  

Among these methods, the MSP \cite{hendrycks2016baseline} is one of the  standard approaches in the current literature. This technique harnesses the output of the softmax function to estimate an OOD probability for each predicted sample.

MLV \cite{MLV} represents a more recent improvement using raw network outputs instead of the normalized ones used by MSP. This has been helpful in cases in which applying the softmax function significantly changes the class scores. An entire dataset can be scored in a single pass through the network.

ODIN \cite{liang2017enhancing} is a simple but effective method that enhances MSP by slightly altering the model workflow. It proposes two different strategies, temperature scaling and input preprocessing that are combined, while the softmax is chosen as the scoring function. Temperature scaling involves dividing the logits by a positive integer constant $T$ before applying the softmax, which increases the gap between in and OOD points.
Input preprocessing, inspired by adversarial attacks, involves adding a small perturbation to a sample $\mathbf{x}$. Each sample is propagated through the network, and the sign of the loss gradient with respect to $\mathbf{x}$ is computed and added to the original point, scaled by a positive constant $\epsilon$. This trick should move input points closer to a peak of the softmax, increasing the gap between those that were already nearby (ID) and the remaining ones (OOD). 
The final score for sample $\textbf{x}$ is given by:

Finally, OODL \cite{oodl} avoids the expensive retraining of the original model, but does not exploit its confidence scores in order to predict the OODness value. Instead, it leverages an external one-class-classifier to decide whether a point is ID or OOD. Aiming to reuse as much as possible the existing model and keep the external classifier simple, the latter does not act on raw samples but rather precomputed features extracted from one of the layers of the original model and then compressed to reduce their dimensionality.

The resulting vector is then fed to the OOD detector, which outputs the final score.

Following the official OODL implementation\footnote{https://github.com/vahdat-ab/OODL} a One-Class SVM can be used as an OOD Detector. This method requires setting a few hyperparameters including, among others, the kernel approximation function (RBF or Nystroem),  the fraction of allowed training errors $\nu$, and the network layer from the features are extracted.

\section{Datasets}
\label{sec:datasets}

In this section, we introduce the proposed benchmarks for OOD detection methods, which include both typical settings from the current OOD literature and a set of novel ones based on a measure of per-class semantic affinity. 

The development of the proposed benchmark stemmed from the \FACETS (Face Aesthetics in Contemporary e-Technological Societies) project, and in particular the FRESCO research line (Face Representations in E–Societies through Computational Observation), which aims at analyzing large collections of profile pictures from social networks through deep learning tools \cite{santangelo2023}. Thus, the proposed benchmark mimics a pretrained image tagging model that needs to operate on social media images, and thus implicitly under an open-world assumption.
Specifically, a scene classification task was selected as the target application, and the Places365 dataset served as the ID set for the benchmark. This choice departs from the mainstream OOD literature, in which ImageNet is typically selected as ID and Places365 as OOD. Although the proposed configuration may be less practical, given that Imagenet pretrained models are more abundant, it also addresses issues that may not emerge when focusing solely on object-centric datasets. For instance scene classifiers may implicitly rely on object detection to perform classification \cite{zhou2014object}, sometimes leading to undesirable outcomes. 

First, we define a \textit{Baseline} dataset that reflects the most basic level of complexity in OOD recognition, based on related works as discussed in Section \ref{sec:related_work}. We chose the validation split of Places365-Standard as the ID data, consisting of 100 labeled images per class from the 365 classes. For the OOD data, we opted for the SVHN in its multi-resolution version due to its semantic difference compared to Places365 and its widespread use in previous studies on similar OOD tasks. To ensure consistency in sample size, we randomly sampled a subset of images from the SVHN test set, resulting in a total of 33,500 OOD images for training the OOD detection \autoref{tab:baseline_ood}.

\begin{table}[tb]

\caption[Composition of the Baseline dataset.]{Composition of the Baseline dataset. SVHN samples are generalized to a single class, e.g. "number", instead of the usual 10 digits. This choice was made because multiple numbers are present in some images. Furthermore, the primary interest in this context is whether OOD samples can be successfully rejected, regardless of the specific digit. }
\label{tab:baseline_ood}

    \centering
    \resizebox{\linewidth}{!}{
    \begin{tabular}{| c | c c c c |} 
         \hline
         \multicolumn{5}{|c|}{\textbf{Baseline Dataset (val/test)}} \\
         \hline\hline
                            & ID classes & OOD classes & ID samples & OOD samples \\
         \hline
         Places365-Standard (val) & 365  &  0          & 18,250     &    0        \\
         SVHN (train)             & 0    &  1          &   0        &  16,701     \\
         SVHN (test)              & 0    &  1          &   0        &  1,549      \\
         \hline
         \textbf{Total}       &\textbf{365}&\textbf{2}&\textbf{18,250}&\textbf{18,250}\\
         \hline
    \end{tabular}
    }

\end{table}

The second dataset used, called \textit{Inter-Dataset OOD Detection}, is inspired by the work by Roady et al.\cite{roady2019out}. In this setting, the objective is to discriminate between two distinct datasets where the OOD samples come from a complex and diverse object-centric dataset, namely ImageNet-1K, which consists of over 1 million training images and 50,000 validation images, covering 1,000 classes. To create a challenging OOD scenario, we removed the 32 common classes between ImageNet-1K and Places365-Standard. The validation split of Places365 was used as the ID dataset, while the OOD samples were collected through stratified random sampling on all remaining classes of ImageNet-1K excluding those present in Places365 \autoref{tab:inter_ood}.

\begin{table}[tb]

\caption[Composition of the splits used for the Inter-Dataset OOD Detection.]{Composition of the  Inter-Dataset benchmark. In this setting the OOD detection task consists of discriminating among the two datasets; therefore, one of them is considered entirely ID and the other as OOD. Common classes were removed.}
\label{tab:inter_ood}
    \centering
    \resizebox{\linewidth}{!}{
        \begin{tabular}{| c | c c c c |} 
         \hline
         \multicolumn{5}{|c|}{\textbf{Inter-Dataset (val/test)}} \\
         \hline\hline
                            & ID classes & OOD classes & ID samples & OOD samples \\
         \hline
         Places365-Standard (val) & 365  &  0          & 18,250     &   0         \\
         ImageNet (train)      &  0   & 968         &   0        & 18,108      \\ 
         \hline
         \textbf{Total}       &\textbf{365}&\textbf{968}&\textbf{18,250}&\textbf{18,108}\\
         \hline
        \end{tabular}
    }
\end{table}

We argue that the standard inter-dataset evaluation protocol, which considers all classes from ImageNet-1K as OOD for a model trained on Places365, is too simplicistic and does not fully capture the underlying relationships between these datasets. Although it is true that there are significant differences between the two datasets in terms of their data collection process, coverage, and label set, some classes from ImageNet-1K and Places365 can be highly correlated from a semantic point of view, and therefore should not consider as separate distributions. For example, an image of an empty room can still be classified as a \textit{bedroom} if the room's design, wall color, and other contextual signs are typical of a bedroom. However, the \textit{bed} object will almost always be visible inside a \textit{bedroom} scenario. By extension, the same picture may be classified as a \textit{bed} in ImageNet and a \textit{bedroom} in Place365, both of which are acceptable predictions. In contrast, in the typical Inter-dataset arrangement, the \textit{bed} class would be classified as OOD and the \textit{bedroom} class as ID. Thus, it is reasonable to consider some classes from ImageNet-1K as part of the ID set when evaluating models on Places365.

One important aspect of selecting a suitable split for comparing models trained on different datasets involves establishing a (ideally) reliable and objective measure of similarity between classes. To achieve this goal, we used the WordNet lexicon \footnote{https://wordnet.princeton.edu/}, which provides a framework for linking classes from various datasets based on semantic relationships.
WordNet organizes nouns, verbs, adjectives, and adverbs into cognitive synonym sets (\emph{synsets}), representing distinct concepts related by various semantic connections.

Computing similarity metrics between ImageNet-1K classes and Places365 classes required establishing a correspondence between these classes and WordNet synsets. While ImageNet class labels are derived from WordNet synsets, Places365 labels consist of words rather than synsets, and hence a single ID class could be associated with more than one WordNet concept. Although much of the mapping was automatic, manual intervention was needed for scene labels that did not fit neatly into a WordNet synset. After obtaining said mapping, we employed a distance metric between semantic concepts.
An average of three path-based metrics, Path (a baseline metric equal to the inverse of the shortest path between two concepts), Leacock-Chodorow \cite{leacock1998combining}, and Wu-Palmer \cite{wu1994verb}, was used as the final similarity metric to account for all with a single score. Different thresholds can then be selected to determine which classes should be considered OOD. The resulting datasets are described in \autoref{tab:wordnet_imagenet_datasets_composition}.

\begin{table}
\caption{Composition of the WordNet ImageNet datasets}
\label{tab:wordnet_imagenet_datasets_composition}
\begin{center}
\resizebox{\linewidth}{!}{
\begin{tabular}{| c | c c c c |} 
 \hline
 \multicolumn{5}{|c|}{\textbf{WordNet ImageNet T40 (val/test)}} \\
 \hline\hline
                    & ID classes & OOD classes & ID samples & OOD samples \\
 \hline
 Places365-Standard (val) & 365  &  0          & 18,250     &   0         \\
 ImageNet (train)      &  56  & 944         &  2,800     & 21,332      \\
                                                         
 \hline
 \textbf{Total}       &\textbf{421}&\textbf{944}&\textbf{21,050}&\textbf{21,332}\\
 \hline
\end{tabular}
}
\resizebox{\linewidth}{!}{
\begin{tabular}{| c | c c c c |} 
 \hline
 \multicolumn{5}{|c|}{\textbf{WordNet ImageNet T45 (val/test)}} \\
 \hline\hline
                    & ID classes & OOD classes & ID samples & OOD samples \\
 \hline
 Places365-Standard (val) & 365  &  0          & 18,250     &   0         \\
 ImageNet (train)      & 90   & 910         & 4,500      & 22,750      \\
                                                         
 \hline
 \textbf{Total}       &\textbf{455}&\textbf{910}&\textbf{22,750}&\textbf{22,750}\\
 \hline
\end{tabular}
}
\resizebox{\linewidth}{!}{
\begin{tabular}{| c | c c c c |} 
 \hline
 \multicolumn{5}{|c|}{\textbf{WordNet ImageNet T50 (val/test)}} \\
 \hline\hline
                    & ID classes & OOD classes & ID samples & OOD samples \\
 \hline
 Places365-Standard (val) & 365  &  0          & 18,250     &   0         \\
 ImageNet (train)      & 140  & 860         & 7,000      & 25,560      \\
                                                         
 \hline
 \textbf{Total}       &\textbf{505}&\textbf{860}&\textbf{25,250}&\textbf{25,560}\\
 \hline
\end{tabular}
}

\end{center}
\end{table}

Lastly, as an alternative to automatic path-based metrics, we manually annotated classes from the the ImageNet and SUN397 \cite{5539970} datasets. Our ultimate goal was to determine whether each class in these datasets should be labeled as ID or OOD relative to the source distribution of Places365. Since the validation split of the Places365-Standard dataset was considered the most reliable reference for ID samples, it was retained for this purpose. On the other hand, SUN397 was included in the design of the last two datasets for both ID and OOD data. It has a strong connection with Places365, with 294 classes shared between the two datasets, and others being similar. The inclusion of SUN397 allowed for the acquisition of additional samples, from a different data source, for ID classes. This choice allowed us to differentiate between the semantic shift due to the occurrence of different classes between OOD and ID samples, from the domain shift that may be introduced by the process of dataset collection. 
However, not all non-overlapping classes can unequivocally be designated as OOD. As a result, manual inspection is necessary to recognize and pair similar classes. We assigned each class an OODness score ranging from 0 to 3 based on the following criteria:
\begin{enumerate}
    \setcounter{enumi}{-1}
    \item classes from Place365 that appear with exact labels.
    \item classes that are semantically related to a class from Place365.
    \item classes that usually contain features characteristic of one or more ID classes.
    \item the remaining classes.
\end{enumerate}

The availability of fine-grained labelling enables experimentation with different dataset configurations. Two main versions of the dataset were investigated (\autoref{tab:facets_t1t2}): T1, in which classes labelled as 0 or 1 are considered ID while 2 and 3 are considered OOD, and T2, where 0, 1 and 2 classes are considered ID and classes labelled as 3 are considered OOD. In contrast, each class in ImageNet-1K is manually given a binary OODness label of 0 or 1, with classes that appear in Places365 or represent objects typically associated with a scene from the target dataset being assigned the label 0 (ID), 1 (OOD) otherwise. The choice of the threshold for the classes drawn from SUN does not affect the way ImageNet classes are considered. 

The selected OOD detection method do not necessarily require the availability of OOD samples during training. However, both ID and OOD samples may be necessary for hyperparameter tuning. Separate validation and test sets for each setup were established to properly assess OOD detection performance and prevent any potential data leakage.

\begin{table}[tb!]
\begin{center}
\caption{Composition of the \FACETS{} OOD detection dataset}
\label{tab:facets_t1t2}
\resizebox{\linewidth}{!}{
\begin{tabular}{| c | c c c c |}
 \hline
 \multicolumn{5}{|c|}{\textbf{\FACETS{} OOD Detection T1 (val/test)}} \\
 \hline\hline
                    & ID classes & OOD classes & ID samples & OOD samples \\
 \hline
 Places365-Standard (val) & 365  &  0          & 18,250     &   0         \\
 SUN397                   & 319  & 78          & 46,851     & 7,530       \\    ImageNet (val)        & 356  & 644         & 8,900      & 16,100      \\
 ImageNet (train)      &  0   & 644         &   0        & 50,232      \\
                                                         
 \hline
 \textbf{Total}       &\textbf{1,040}&\textbf{1,366}&\textbf{74,001}&\textbf{73,862}\\
 \hline
\end{tabular}
}
\resizebox{\linewidth}{!}{
\begin{tabular}{| c | c c c c |} 
 \hline
 \multicolumn{5}{|c|}{\textbf{\FACETS{} OOD Detection T2 (val/test)}} \\
 \hline\hline
                    & ID classes & OOD classes & ID samples & OOD samples \\
  \hline
 Places365-Standard (val) & 365  &  0          & 18,250     &   0         \\
 SUN397                   & 351  & 46          & 49,827     & 4,554       \\    ImageNet (val)        & 356  & 644         & 8,900      & 16,100      \\
 ImageNet (train)      &  0   & 644         &   0        & 56,606      \\
                                                         
 \hline
 \textbf{Total}       &\textbf{1,072}&\textbf{1,334}&\textbf{76,977}&\textbf{77,260}\\
 \hline

\end{tabular}
}

\end{center}
\end{table}

\section{Preliminary Analysis}
\label{sec:preliminary_analysis}

The primary focus of OOD detection should not merely revolve around the ability to distinguish images drawn from ImageNet-1K or Place365, but to discriminate inaccurate or deceptive outputs compared to plausible predictions, irrespective of the specific set of labels used for dataset annotation. As an initial step to obtain a better understanding of the behavior of a typical scene classifier (Resnet50 pre-trained on Places365), an analysis of network predictions was performed. The objective was to ascertain whether any consistent patterns of misclassification emerged. This analysis was carried out on the \FACETS OOD Detection T1 dataset.

Although the prevalent metric for evaluating Places365 is top-5 accuracy, a different approach is necessary to evaluate predictions on OOD samples due to the mismatch in class labels. The present analysis addresses the question of \textit{"how closely does the predicted class align with a correct one?"} To tackle this question, a strategy similar to the labeling methodology employed for the WordNet-ImageNet datasets is adopted, utilizing the semantic similarity between class labels.

Similarly to the mapping technique elucidated in Section~\ref{sec:datasets}, both the SVHN and SUN397 datasets were mapped to WordNet synsets to ensure methodological uniformity. For SVHN a manual mapping to \textit{digit.n.01} was performed. Regarding SUN397, the mappings were conducted partially automatically, employing the same procedure adopted for Places365. The significant overlap in classes between the two datasets facilitated this process, allowing reuse of 294 out of the 397 required mappings, since these classes are shared with Places365. The similarity score between the ground truth class and the predicted class was computed as the average of the Wu-Palmer and the Path similarity metrics. Other similarity metrics were disregarded due to their unbounded nature, which could lead to dominate those bounded between $0$ and $1$.

\begin{figure}[tb!]
    \centering
    \includegraphics[width=\columnwidth]{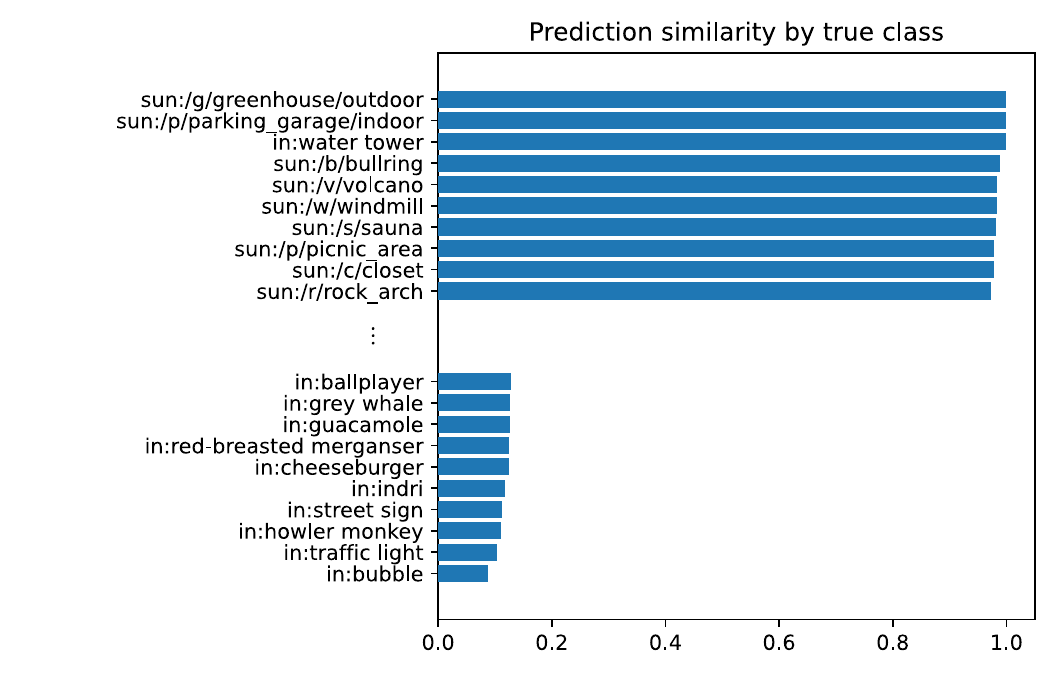}
    \caption[Average semantic similarity between the ground truth category and the predicted ID labels.]{Average semantic similarity between the ground truth class and the predicted ID labels, as computed on the validation split of the \FACETS{} OOD Detection T1. SUN ground truth labels (\textit{sun} prefix) were generally semantically similar to the predicted ID class if compared to ImageNet ground truth labels (\textit{in} prefix).}
    \label{fig:FACETST1_pred_sim_by_gt}
\end{figure}

The same scoring function was utilized to perform aggregated measurements to assess the behavior of the pre-trained model. \autoref{fig:FACETST1_pred_sim_by_gt} presents the average similarity between the ground truth labels of each sample from the original dataset and the predicted classes from the classifier trained on Places365.  It should be noted that none of the classes in the Places365 validation set are included among the top 10 best performing classes, indicating a substantial semantic similarity between the SUN and Places365 classes. On the other hand, ImageNet classes generally exhibit poorer performance because of the typically limited semantic similarity between objects and their backgrounds.

\begin{figure}[ht!]
    \centering
    \includegraphics[width=\columnwidth]{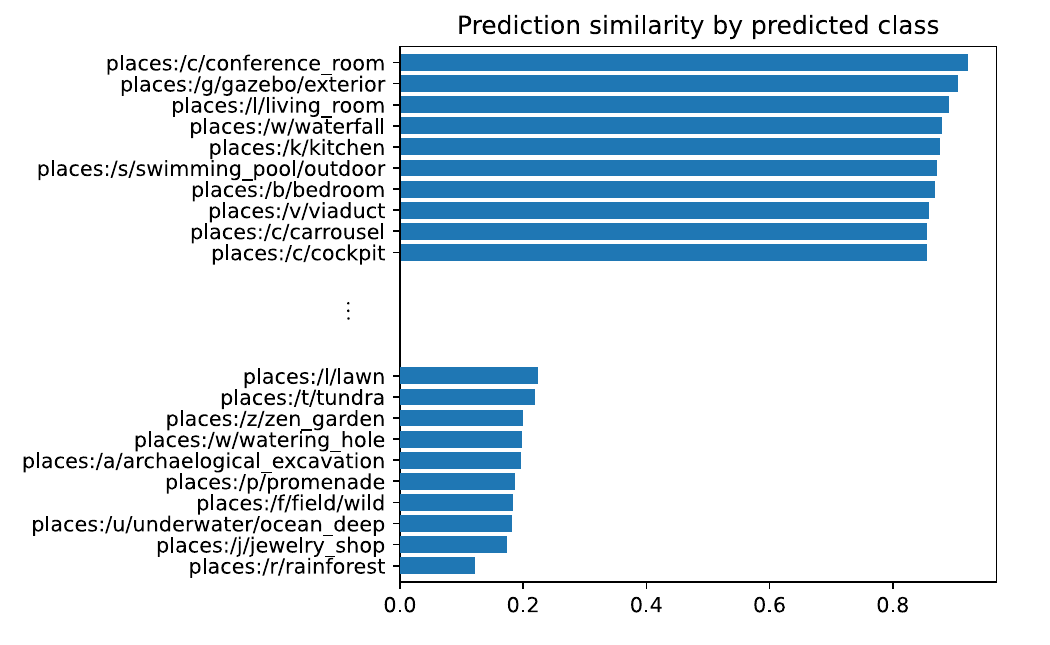}
    \caption[Average semantic similarity between the ground truth category and the predicted ID labels.]{Average semantic similarity between the OOD ground truth class and the predicted ID labels, as computed on the validation split of the \FACETS{} OOD Detection T1. Man-made environments seems to be less ambiguous and predicted classes are most likely to be semantically similar to the respective OOD ground truth class.}
    \label{fig:FACETST1_pred_sim_by_pred}
\end{figure}

Similarly, \autoref{fig:FACETST1_pred_sim_by_pred} showcases the 10 best/worst ID classes (derived from the Places365 dataset) based on the average similarity between the predicted ID label and the ground truth. For example, the \textit{conference\_room} class from Places365 exhibits a high similarity with the ground truth labels of the samples assigned to that class, indicating that the model assigns a category that is semantically close to the actual class for OOD samples. In general, it appears that indoor and man-made environments are more prone to receive plausible labels, whereas the same does not hold for natural scenes. Various explanations can account for this phenomenon: artificial settings tend to have less ambiguity in terms of labeling compared to landscapes, where multiple classes may be present, such as hills, skies, or forests.

To explore the network's behavior beyond the previously mentioned similarity metric, an alternative visualization method involves representing relationships as a directed graph. Each class is depicted as an individual node, and the presence of an edge from node $a$ to node $b$ indicates that at least one image of class $a$ was predicted as class $b$. The resulting graph structure enables the identification of overarching trends in the classification process. To incorporate the frequency and significance of each association, the weight of each edge in the graph shows the strength of the connection between the respective nodes. 

The complete graph for \FACETS OOD Detection T1, established by utilizing MSP as the OOD detection score, is quite large, encompassing $1,760$ nodes and $41,397$ edges.  Several pruning procedures were implemented to eliminate redundant and noisy information. Self-loops, which represent accurate classifications but do not contribute to OOD detection insights, were removed. Similarly, links between evidently ID samples were discarded. Associations between classes sharing identical names, such as \textit{sun:/f/fountain}, \textit{in:fountain}, and \textit{places:/f/fountain}, were also removed.  Furthermore, edges originating from Places365 (val) were eliminated, resulting in the remaining edges connecting classes from different datasets. Edges with low weights, which likely offered minimal informative value while complicating the graph's structure, were also pruned. The nodes with a resulting degree of zero were finally removed. The pruned graph has a more manageable scale, consisting of 989 nodes and 1022 edges. To enable easier navigation, the graph will be released along with the dataset. 

\begin{figure}
    \centering
    \includegraphics[width=\columnwidth]{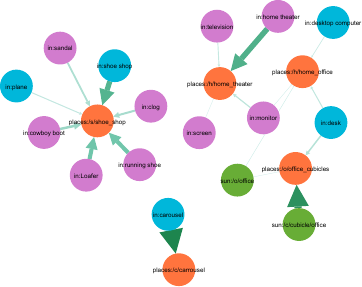}
    \caption[Examples of strong edges between classes representing the same concepts with slightly different names.]{Examples of strong edges between classes representing the same concepts with slightly different names. The underscore prevented \textit{shoe shop} and \textit{home theater} to be paired with their counterparts, whereas different wording or spelling were responsible for mismatches in the case of \textit{cubicle/office} and \textit{carrousel}.}
    \label{fig:graph_class_names}
\end{figure}

Within this refined representation, distinctive isolated clusters that suggest the presence of independent semantically related groups were identified.  Furthermore, numerous high-weight edges were observed, highlighting connections that were overlooked by the basic algorithm employed for class name matching. It should be noted that subtle variations such as the presence of an underscore instead of a white space or a different word order were sufficient to classify two classes as distinct entities (\autoref{fig:graph_class_names}). 

Several other clusters exemplify meronym-holonym or scene-object relationships, which are not adequately captured by widely used WordNet similarity metrics. As an  example, a prominent cluster comprises classes representing animals. Notably, the nodes that attract a significant number of connections are \textit{veterinarian\_office} and \textit{underwater/ocean\_deep}, indicating terrestrial and aquatic species, respectively. Other classes frequently assigned as outputs for animal classification include \textit{aquarium}, \textit{field/wild}, \textit{watering\_hole}, \textit{lawn}, \textit{kennel/outdoor}, \textit{pet\_shop}, \textit{tundra}, and \textit{rainforest}. Although these scenes are generally acceptable for animals, and the species often align with their habitats, the strength of certain associations suggests potential network biases. For instance, the fact that most dogs and domestic species are associated with \textit{veterinarian\_office} or \textit{kennel}, instead of their typical environments such as a house or a garden, implies that the model may classify scenes with pets without adequately considering the background. This bias could be attributed to the significantly higher occurrence of pets in the \textit{veterinarian\_office} class compared to other classes in Places365. Finally, undesirable behaviour emerge from clusters of unrelated concepts, that are very likely originated by visual rather than semantic similarity: for instance, towers, lighthouses and telescopes are all classified  \textit{bank\_vault}  due to their round shape.

\section{Results}
\label{results}
The effectiveness of the chosen OOD detection methods is shown in \autoref{tab:ood_detection_results}, and the best score for each dataset is highlighted in bold. The differences across datasets are evident: most approaches are successful when dealing with the Baseline, but struggle when the complexity increases. Moving to InterDataset OOD Detection causes the first noticeable drop as the ID and OOD distribution become closer. As observed by the model's worse performance on the WordNet ImageNet datasets, the requirement for discrimination based on the semantic content also impairs its capacity to identify outliers. That becomes even more evident when more classes and data sources are included, such as \FACETS OOD Detection datasets. The drop in performance is particularly evident for OODL, which among the tested approaches is the only one that relies on a decision function explicitly trained on each dataset. 

\begin{table*}[ht]
   \caption[OOD detection results for each dataset.]{OOD detection results: the best score is highlighted in bold. Specifically, for each dataset, the OOD detection techniques that score the smallest FPR@95\%TPR, the smallest detection error, and the largest AUROC are highlighted. If multiple techniques have equal performance, both scores are highlighted in bold.}
    \label{tab:ood_detection_results}
    \centering
    \renewcommand{\arraystretch}{1.5}
    \resizebox{0.9\textwidth}{!}{
    \begin{tabular}{|c|c|c|c|c|c|c|}
        \hline
                                   & \multicolumn{6}{c|}{FPR@95\%TPR $\downarrow$ / Detection Error $\downarrow$ / AUROC $\uparrow$} \\
        \hline
                                   & MSP \cite{hendrycks2016baseline}              & TS \cite{liang2017enhancing}                & MLV \cite{MLV}              & ODIN \cite{liang2017enhancing}              & IP TS MLV \cite{liang2017enhancing}        & OODL \cite{oodl}          \\
        \hline
        Baseline                   & 22.36/13.68/94.79 &  4.50/ 4.75/99.02 &  4.50/ 4.74/99.02 & 10.34/ 7.67/97.82 & 10.34/ 7.67/97.82 &  \textbf{0.37}/ \textbf{1.61}/\textbf{99.81} \\
        InterDataset OOD Detection & 86.85/45.91/64.40 & 79.69/42.34/69.79 & \textbf{79.67}/\textbf{42.33}/69.79 & 84.43/44.72/65.73 & 84.46/44.72/65.74 & 80.83/42.92/\textbf{71.30} \\
        WordNet ImageNet T40       & 87.22/46.11/63.99 & 81.28/\textbf{43.14}/68.91 & \textbf{81.27}/\textbf{43.14}/68.91 & 85.27/45.13/64.84 & 85.26/45.13/64.85 & 82.06/43.53/\textbf{70.93} \\
        WordNet ImageNet T45       & 87.64/46.32/63.44 & 82.16/\textbf{43.58}/68.05 & \textbf{82.15}/\textbf{43.58}/68.05 & 86.44/45.72/63.80 & 86.43/45.71/63.81 & 83.19/44.09/\textbf{70.24} \\
        WordNet ImageNet T50       & 88.66/46.83/62.28 & \textbf{85.30}/\textbf{45.15}/66.29 & \textbf{85.30}/\textbf{45.15}/66.29 & 88.14/46.56/62.30 & 88.14/46.56/62.30 & 85.37/45.18/\textbf{69.54} \\
        \FACETS{} OOD Detection T1    & 82.64/43.82/69.56 & \textbf{74.19}/39.60/\textbf{75.87} & \textbf{74.19}/\textbf{39.59}/\textbf{75.87} & 79.11/42.05/72.39 & 79.10/42.05/72.40 & 82.24/43.62/67.49 \\
        \FACETS{} OOD Detection T2    & 82.44/43.72/69.83 & \textbf{74.01}/\textbf{39.51}/\textbf{76.38} & 74.02/\textbf{39.51}/\textbf{76.38} & 78.92/41.96/72.70 & 78.93/41.96/72.70 & 81.49/43.25/68.43 \\
        \hline
    \end{tabular}}
 
\end{table*}

\begin{table}[ht]
\caption{Misclassification detection results on Places365-Standard (val)}
    \label{tab:misclassification_detection_results}
    \centering
    \resizebox{\linewidth}{!}{
    \begin{tabular}{|c|c|c|c|c|c|}
        \hline
                                    \multicolumn{6}{|c|}{FPR@95\%TPR $\downarrow$ / Detection Error $\downarrow$ / AUROC $\uparrow$} \\
        \hline
                                    MSP \cite{hendrycks2016baseline}               & TS \cite{liang2017enhancing}                 & MLV \cite{MLV}              & ODIN \cite{liang2017enhancing}             & IP TS MLV \cite{liang2017enhancing}           & OODL \cite{oodl}              \\
        \hline
                                    \textbf{76.33}/\textbf{40.66}/\textbf{77.74} & 81.92/43.45/70.84 & 81.91/43.45/70.84 & 81.83/43.41/71.23 & 81.84/43.41/71.23 & 96.40/49.97/47.24 \\
        \hline
    \end{tabular}}
    
\end{table}

Although MSP is effective as a baseline, more advanced methods often deliver better results. Techniques based on input perturbation, specifically, ODIN and IP TS MLV fail to outperform their baseline alternatives, TS and MLV. In terms of AUROC, OODL appears to be a better choice for less complex datasets. However, classifier-based measures like TS and MLV generally surpass SVM on the \FACETS OOD Detection datasets, achieving higher scores even on simpler benchmarks like WordNet and ImageNet. 

The difference with respect to the adopted method could be due to the difficulties faced by SVMs when the required decision boundary becomes very complex, while the one with respect to WordNet datasets is likely to be a consequence of the manual labelling: it is possible that some of the labels pander to existing network biases, rewarding it for its mistakes rather than applying penalties. such as incorrect labeling of certain animal categories in ImageNet. These errors can lead to false positives in OOD detections, where test images featuring these animals in unfamiliar settings were incorrectly labeled as ID but received high scores from softmax and logit-based methods. This misclassification inflates performance metrics but is an outcome of both the ground truth label and classifier output being incorrect.

The last aspect investigated was the dependability and robustness of the OOD detection metric in identifying misclassified examples in the ID dataset \autoref{tab:misclassification_detection_results}. This is crucial as it affects how well the approach can identify incorrect classification results. Although MSP did not perform optimally in OOD detection, it emerged as the best option among those tested in terms of identifying misclassified samples.  In this context, the softmax approach gains an advantage due to the absence of unfamiliar classes, consistent with the key "closed world assumption". During the normalization process, the total probability of the output classes must equal 1, resulting in higher confidence levels for correctly classified instances while reducing the ability to detect outliers. Outlier detection techniques such as OODL do not achieve high performance in this task, as expected given the way in which the additional classifier is trained on the ID distribution. 

\section{Conclusion}
\label{sec:conclusions}
With a focus on the semantic content of the images, the present study focuses on the problem of OOD identification in neural networks by considering as the main case study a scene CNN classifier that was pre-trained on the Places365 dataset. Several OOD detection techniques were analyzed. However, in contrast to previous experimental comparison, we sought to design a benchmark that better reflects practical applications. Consequently, instead of popular toy datasets like CIFAR10, CIFAR100, or TinyImageNet, we leveraged higher resolution images from ImageNet and Places365 to design a more realistic and challenging benchmark comprising both far-OOD and near-OOD samples. Similar to previous studies \cite{semantic_anomalies,semantically_coherent_oodd}, our benchmark emphasizes OOD classification as a measure of semantic similarity by assigning each individual class as ID or OOD based on the semantic content, rather than simply rely on the dataset of origin. To this aim, we compared manual and automatic labelling. For the latter, we leveraged the WordNet database and computed similarity based on concepts, resulting in the creation of three WordNet-ImageNet datasets. 

Several experimental decisions were guided by practical considerations or limitations. For example, our choice to concentrate on OOD detection approaches that did not necessitate re-training the classifier was prompted by the availability of pre-trained models provided by third-party sources. Hence, there is ample room for further advancements. Using a single ID distribution and model is a limitation of the current study. Likewise, the present study could be extended by experimenting with additional OOD detection strategies and investigating an ensemble of metrics.

During the development of the datasets, efforts were made to create a finer OOD label assignment, in which each class, rather than dataset, was designated as ID or OOD. Although this is a positive step forward, it is still far from being optimal, especially when handling object-centric datasets in which images belonging to the same category may depict vastly differing scenes, requiring some to be categorized as ID and others as OOD. Manual inspection of every image is impractical, hence the operation must be automated. At the class level, enhancing the semantic distance metric would also improve the labeling of the WordNet-ImageNet datasets, and facilitate a more accurate comparison between concept proximity and output OOD scores. 

\section{Acknowledgements}
This project has received funding from the European Union (ERC project FACETS - Face Aesthetics in Contemporary E-Technological Societies, grant agreement 819649). Views and opinions expressed are however those of the author(s) only and do not necessarily reflect those of the European Union or the European Research Council Executive Agency. Neither the European Union nor the granting authority can be held responsible for them.

\printbibliography

\end{document}